%% file: main.tex
\pdfoutput=1
\documentclass[11pt]{article}
\usepackage{acl}

\input{preamble/preamble}
\input{preamble/math_commands}

\input{preamble/notation}

\input{preamble/bibstyle}

\usepackage{tipa}
\newcommand{\ethz}{\boldsymbol{\texttt{1}}}
\newcommand{\copenhagen}{\boldsymbol{\texttt{2}}}
\newcommand{\epfl}{\boldsymbol{\texttt{3}}}
\newcommand{\chicago}{\boldsymbol{\texttt{4}}}

\usepackage{times}
\usepackage{latexsym}

\usepackage[T1]{fontenc}

\usepackage[utf8]{inputenc}
\usepackage{microtype}
\usepackage{inconsolata}

\usepackage{tabularx}
\usepackage{booktabs}

\title{Activation Scaling for Steering and Interpreting Language Models}%

\author{
Niklas Stoehr$^{\ethz}$~\;~
Kevin Du$^{\ethz}$~\;~
Vésteinn Snæbjarnarson$^{\copenhagen}$~\;~ 
\\
\textbf{Robert West$^{\epfl}$}~\;~
\textbf{Ryan Cotterell$^{\ethz}$}~\;~
\textbf{Aaron Schein$^{\chicago}$}
\\
$^{\ethz}$ETH Z{\"u}rich \quad $^{\copenhagen}$University of Copenhagen \quad $^{\epfl}$EPFL \quad $^{\chicago}$The University of Chicago
\\
\href{mailto:niklas.stoehr@inf.ethz.ch}{\texttt{niklas.stoehr@inf.ethz.ch}}~\;~
\href{mailto:kevidu@ethz.ch}{\texttt{kevin.du@inf.ethz.ch}}~\;~ 
\href{mailto:vesteinn.snaebjarnarson@gmail.com}{\texttt{vesn@di.ku.dk }}\\
\href{mailto:robert.west@epfl.ch}{\texttt{robert.west@epfl.ch}}~\;~ 
\href{mailto:ryan.cotterell@inf.ethz.ch}{\texttt{ryan.cotterell@inf.ethz.ch}}~\;~ 
\href{mailto:schein@uchicago.edu}{\texttt{schein@uchicago.edu}}
}

\begin{document}
 \maketitle

\begin{abstract}
Given the prompt \setQuote{Rome is in}, can we steer a language model to flip its prediction of an incorrect token \setQuote{France} to a correct token \setQuote{Italy} by only multiplying a few relevant activation vectors with scalars? We argue that successfully intervening on a model is a prerequisite for interpreting its internal workings. Concretely, we establish a three-term objective: a successful intervention should flip the correct with the wrong token and vice versa (effectiveness), and leave other tokens unaffected (faithfulness), all while being sparse (minimality). Using gradient-based optimization, this objective lets us learn (and later evaluate) a specific kind of efficient and interpretable intervention: activation scaling only modifies the signed magnitude of activation vectors to strengthen, weaken, or reverse the steering directions already encoded in the model. On synthetic tasks, this intervention performs comparably with steering vectors in terms of effectiveness and faithfulness, but is much more minimal allowing us to pinpoint interpretable model components. We evaluate activation scaling from different angles, compare performance on different datasets, and make activation scalars a learnable function of the activation vectors themselves to generalize to varying-length prompts.\footnote{Code to experiment with our method is available at \href{https://github.com/niklasstoehr/activationScaling}{\color{darkblue} \footnotesize \texttt{https://github.com/niklasstoehr/activationScaling}}.}

\end{abstract}

\begin{figure}[t!]
 \centering
 \includegraphics[width=1.0\linewidth]{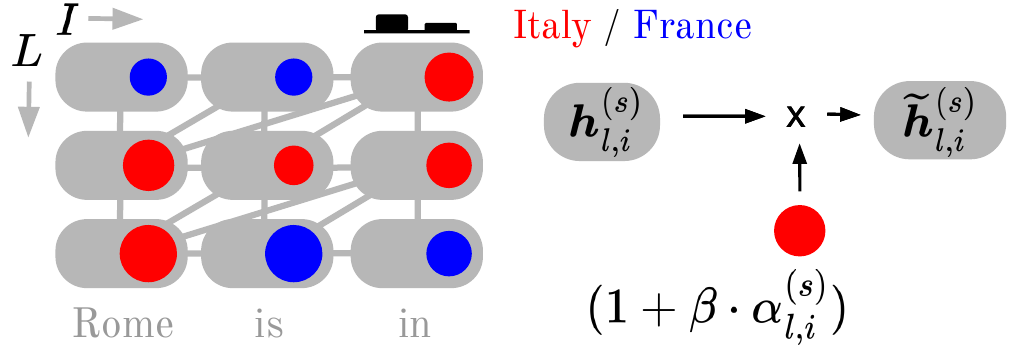} 
\caption{We show that it is often sufficient to scale a few influential activation vectors $\activs{\layer, \tokenPos}{(\site)}$ for a model to favor one answer token over another token. This could be the MLP output at layer $\layer$ for token position $\tokenPos$. We learn multiplicative scalars, $\scalar{(\site)}$, using gradient-based optimization. These correspond to interpretable interventions that generalize to test set prompts while relying on fewer parameters than additive steering vectors.
}
\label{fig:overview}
\end{figure}

\section{Introduction}
\label{sec:intro}

Understanding which components of a language model play which roles in which tasks is a core aim of mechanistic interpretability. Given the prompt \promptStr{Rome is in}, for instance, one might ask which components of the model most influence it to favor \correctAnsw{Italy} over some incorrect answer token, such as \wrongAnsw{France}.
In addressing this question, a natural axiom is that a given component can only be understood as influential for a given task if intervening on it meaningfully alters the model's task-specific behavior.\looseness=-1

Building on this basic axiom, a growing literature seeks to both generate and test hypotheses about where certain behaviors are localized in a model by employing targeted interventions with methods such as activation patching \citep{lakretz_emergence_2019, vig_causal_2020, meng_locating_2022}, among others. 
Studies in this literature often produce a set of attribution scores associated with various locations in the model which represent how much the model's output changed after editing the activation vectors at each location.
Although an effective intervention may be necessary to believe a given localization hypothesis, it is not sufficient, as interventions to other model locations may be similarly effective. 
Indeed, recent work has questioned the relationship between interpretability and intervention on this basis, and has advocated for more rigorous and deliberate methodology for connecting the two~\citep{hase_does_2023, wang_does_2024, hanna_have_2024}.\looseness=-1

A parallel literature on model steerability also seeks to develop effective interventions, not for the primary purpose of interpretability, but to steer models toward desirable behaviors, like factuality~\citep{li_inference-time_2023}, or away from undesirable behaviors, like toxicity~\citep{ilharco_editing_2023, turner_activation_2023}. 
Methods in this literature are typically designed to be maximally effective for steering and thus tend to involve model-wide (i.e., not necessarily localized) interventions to the weights \citep{houlsby_parameter-efficient_2019, hu_lora_2021, ilharco_editing_2023} or activation vectors \citep{subramani_extracting_2022, hernandez_inspecting_2024}.~\looseness=-1

This paper seeks to synthesize the goals of model-wide interventions while still offering mechanistic insights into the model. Specifically, we seek interventions that are \defn{effective}, and not localized \textit{a priori}, but are nevertheless \defn{minimal} and \defn{faithful}. We define an intervention to be effective if it flips the prediction of the \correctAnsw{correct} answer token (e.g., \correctAnsw{Italy}) to an \wrongAnsw{incorrect} token (e.g., \wrongAnsw{France}). We define an intervention to be faithful if it does not substantially alter the probabilities of tokens unrelated to the given task. And, we define an intervention to be minimal if it alters activation vectors sparsely, in only a small number of locations. Our notions of minimality and faithfulness follow from recent work in the Transformer circuits literature \citep{wang_interpretability_2023, bhaskar_finding_2024}. We operationalize these three desiderata via a three-term objective allowing us to learn model-wide interventions using gradient-based optimization.~\looseness=-1

More specifically, we propose a kind of model-wide yet parsimonious intervention that associates a single scalar parameter with each of many locations in a model. The scalar parameters are learned so that the model can be effectively steered---e.g., away from \correctAnsw{Italy} and toward \wrongAnsw{France}---by simply scaling the activation vector at each location. We call this approach \defn{activation scaling} (\FixedAttr), inspired by the idea that some model components are highly specialized for certain task-relevant computations \citep{voita_analyzing_2019, geva_transformer_2022}. Existing work \citep{yu_characterizing_2023, merullo_circuit_2024, ortu_competition_2024} has identified individual components, e.g., name mover heads \citep{wang_interpretability_2023}, in a first step and successfully steered a model by scaling only the contributions of those individual components in a second step.

To evaluate our approach, we construct a baseline method that is the same in all aspects except it learns additive vectors at all locations, rather than single multiplicative scalars. We refer to this approach as \defn{steering vectors} (\FixedVec). In a suite of experiments, we find overall that activation scaling learns interventions which are as effective and faithful as steering vectors, while requiring far fewer learnable parameters. Our results suggest that merely scaling the signed magnitude of activation vectors, without further affecting their direction, is sufficient for effective steering on simple tasks. Moreover, we find activation scalars are highly interpretable. They are easy to understand as simply strengthening or weakening the steering directions already encoded in the model \citep{subramani_extracting_2022, ferrando_information_2024}, and when visualized, they provide sparse and localized descriptions of important model components. Finally, to extend our approach beyond fixed prompt templates, we develop a dynamic version of activation scaling. \DynamicAttr makes the activation scalars learned functions of the activation vectors themselves, thus allowing learned interventions to transfer to test set prompts of varying length.\looseness=-1

\begin{table*}[t]
\fontsize{9}{9}\selectfont
\centering
\renewcommand{\arraystretch}{1.4} 
\setlength{\tabcolsep}{0.35em} 
\begin{tabularx}{\textwidth}{lXl}
\toprule
\textbf{Dataset} & \textbf{Prompt} & \textbf{Answer Toks} \\ 
\midrule
Country--Capital Conflict (\CCC) & \promptStr{The capital of Germany is Paris. Q: What is the capital of Germany? A:} & \correctAnsw{Berlin} / \wrongAnsw{Paris} \\
Indirect Object Identification (\IOI) & \promptStr{When Anne met with Tom, Tom gave the book to} & \correctAnsw{Anne} / \wrongAnsw{Tom}\\
\bottomrule
\end{tabularx}
\caption{We study two datasets commonly used for mechanistic interpretability: conflicts (\CCC) between correct and incorrect facts about country capitals; Indirect Object Identification (\IOI) requiring weak syntactic reasoning.}
\label{tab:datasets}
\end{table*}

\section{Transformer Language Models}
\label{sec:background}

Let $\Sigma$ be an alphabet of \defn{tokens}, a finite, non-empty set, and let $\SigmaStar$ be the set of all strings with tokens drawn from $\Sigma$. A \defn{language model} $\prob$ is a probability distribution over $\SigmaStar$. As is current practice, most language models are defined autoregressively.
Let $\String = \stringtok_{1} \cdots \stringtok_{I} \in \SigmaStar$ be a string; then the autoregressive factorization of $\prob$ is given by
\begin{equation}
    \prob(\String) = \prob(\EOS \mid \String) \prod_{\tokenPos=1}^{I} \prob(\stringtok_{\tokenPos} \mid \String_{< \tokenPos}).
\end{equation}
Each local conditional distribution $\prob(\cdot \mid \String_{<\tokenPos})$ is a distribution over $\SigmaBar = \Sigma \cup \EOS$, where $\EOS \not\in \Sigma$ is the end-of-string token.
In the context of an autoregressive language model, we call $\String_{<\tokenPos}$ a \defn{prompt}.\looseness=-1

Let $\prompt \in \SigmaStar$ be a prompt of length $\TokenPos$ and $\sy \in \SigmaBar$ the next token. A common way to define the local conditional is via the softmax function $\sigma$ which maps from $\mathbb{R}^{|\SigmaBar|}$ to the probability simplex $\Delta^{|\SigmaBar|-1}$:~\looseness=-1
\begin{align}
   \prob(\sy \mid \prompt)
   &= \sigma(\model(\prompt))_{\sy} \nonumber \\ &= \frac{\exp\big(\model(\prompt) _{\sy}\big)}{\sum_{y' \in \overline{\Sigma}} \, \exp\big(\model(\prompt)_{y'}\big)},
\end{align}
where $\model(\prompt) = \Unembed \activs{\Layers, \TokenPos}{}$ defines the \defn{logit function} $\model \colon \SigmaStar \rightarrow \mathbb{R}^{|\SigmaBar|}$ of a language model, where $\Unembed \in \mathbb{R}^{|\SigmaBar| \times D}$ is the projection (or unembedding) matrix, and where $\activs{\Layers, \TokenPos}{} \in \mathbb{R}^{D}$ is the \defn{activation vector} at the final model layer $\Layers$ and last token position $\TokenPos$ of the prompt.\looseness=-1

Most state-of-the-art language models rely on the Transformer architecture \citep{vaswani_attention_2017} to compute $\model$. Transformers are composed of $\Layers$ layers of Transformer blocks, each of which consists of a multi-headed attention $\ATTN{\layer}$ and a multi-layer perceptron $\MLP{\layer}$ function that read from and write into the residual stream. For instance, a Transformer block in a \gpt \citep{radford_language_2019} or \pythia \citep{biderman_pythia_2023} model is given by
\begin{subequations}
\begin{align}
\Activs{\layer}{(1)} &\overset{}{=} \ATTN{\layer} \big( \layerNorm{\layer}{(1)}(\Activs{\layer-1}{(4)}) \big)\\
\activs{\layer, \tokenPos}{(2)} &\overset{}{=} \activs{\layer, \tokenPos}{(1)} + \activs{\layer-1, \tokenPos}{(4)}\\
\activs{\layer, \tokenPos}{(3)} &\overset{}{=} \MLP{\layer} \big( \layerNorm{l}{(2)}(\activs{\layer, \tokenPos}{(2)}) \big)\\
\activs{\layer, \tokenPos}{(4)} &\overset{}{=} \activs{\layer, \tokenPos}{(3)} + \activs{\layer, \tokenPos}{(2)}\\
\Activs{\layer}{(4)} &\overset{}{=} [ \activs{\layer, 1}{(4)}, \ldots, \activs{\layer, \TokenPos}{(4)} ],
\end{align}
\label{eq:transformer_block}%
\end{subequations}%
where $\activs{\layer, \tokenPos}{(\site)} \in \mathbb{R}^{D}$ is an activation (column) vector at a specific \defn{site} $\site$ at \defn{layer} $\layer$ and \defn{token position} $\tokenPos$, and $\layerNorm{\layer}{(\site)}$ is the pre-layer normalization \citep{ba_layer_2016, xiong_layer_2020}. 
We stack activation vectors across token positions to obtain a layer-wise activation matrix $\Activs{\layer}{(\site)} \in \mathbb{R}^{D \times \TokenPos}$. 
The matrix $\Activs{1}{(4)}$ is initialized to a representation of the input $\prompt$.\looseness=-1

\section{Activation-Level Interventions}
\label{sec:method}

\subsection{Choosing Intervention Points}

We focus on a class of interventions which modify one or more of the activation vectors in \cref{eq:transformer_block}.\footnote{We refer to selected sites in \cref{eq:transformer_block} with names: \attnOut for $\site = 1$, \mlpOut for $\site = 3$ and \residPost for $\site = 4$.} This level of abstraction is motivated by our desire to interpret larger components of the Transformer.
However, we note that the granularity at which we seek to intervene and understand the model is a choice which depends on specific use cases.
Our intervention targets a set of layer indices $\layerSet$, token positions $\tokenSet$, and sites $\siteSet$. 
We denote the Cartesian product of the layer indices, token positions and sites, $\Points = \layerSet \times \tokenSet \times \siteSet$, as the \defn{intervention points}.  

\subsection{Defining an Intervention}

We intervene on the activation vectors of the model $\model(\prompt)$ by specifying an intervention $\intervmodel(\prompt)$ that involves a set of learnable parameters $\Params$ and a hyperparameter $\beta \in \mathbb{R}$ which controls the strength and direction of the intervention. For instance, setting $\beta = 0$ removes an intervention, resulting in $\intervmodel(\prompt) = \model(\prompt)$, while switching the sign of $\beta$ reverses the intervention's direction. We compute $\intervmodel(\prompt)$ via $\changedActivs{\layer, \tokenPos}{(\site)}$ as follows.

\paragraph{Additive Vectors.}
We first consider an intervention based on steering vectors (\FixedVec). Specifically, we define a set of intervention parameters $\Params = \{ \vecto{(\site)} \}_{(\layer, \tokenPos, \site) \in \Points}$ that associate a vector $\vecto{(\site)}$ with each intervention point. The intervention adds this vector to its corresponding activation vector:
\begin{align}
    \changedActivs{\layer, \tokenPos}{(\site)}= \activs{\layer, \tokenPos}{(\site)} + \beta \vecto{(\site)}. 
    \label{eq:vectors}
\end{align}

\paragraph{Multiplicative Scalars.}

Applying an intervention vector modifies both the direction and magnitude of the activation vector. We instead propose a more parameter-efficient intervention which is restricted to scaling the signed magnitude of each activation vector via a single multiplicative scalar. This approach, which we call activation scalars (\FixedAttr), is given by
\begin{align}
    \changedActivs{\layer, \tokenPos}{(\site)}= \activs{\layer, \tokenPos}{(\site)}( 1 + \beta \scalar{(\site)}).
    \label{eq:scalars}
\end{align}
where $\Params =  \{ \scalar{(\site)} \}_{(\layer, \tokenPos, \site) \in \Points}$ are the parameters.
\subsection{Learning an Intervention}
\label{sec:objective}

What qualities does an interpretable intervention possess? In this article, we focus on interventions that are \defn{effective}, \defn{faithful}, and \defn{minimal}, drawing on analogous concepts established in the Transformer circuits literature
\citep{wang_interpretability_2023, bhaskar_finding_2024}.

\begin{figure*}[t!]
 \centering
    \raisebox{0.6in}{\rotatebox[origin=t]{90}{\CCC}}  
    \begin{subfigure}[b]{0.48\textwidth}
       \includegraphics[width=\textwidth]{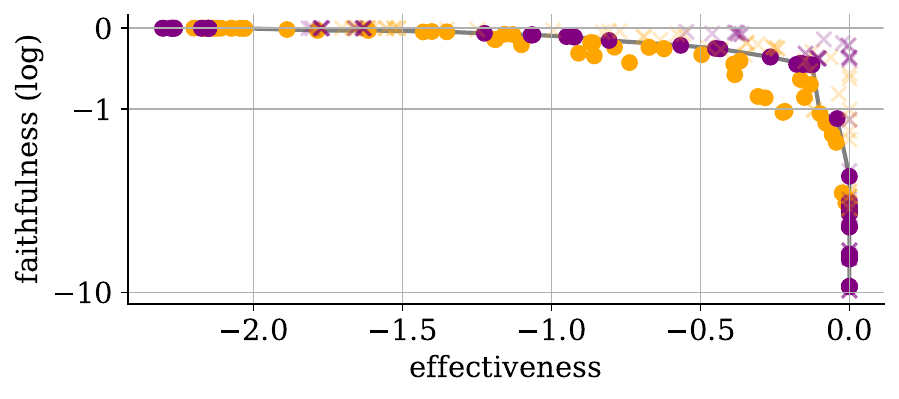}
   \end{subfigure}\hfill
    \begin{subfigure}[b]{0.48\textwidth}
       \includegraphics[width=\textwidth]{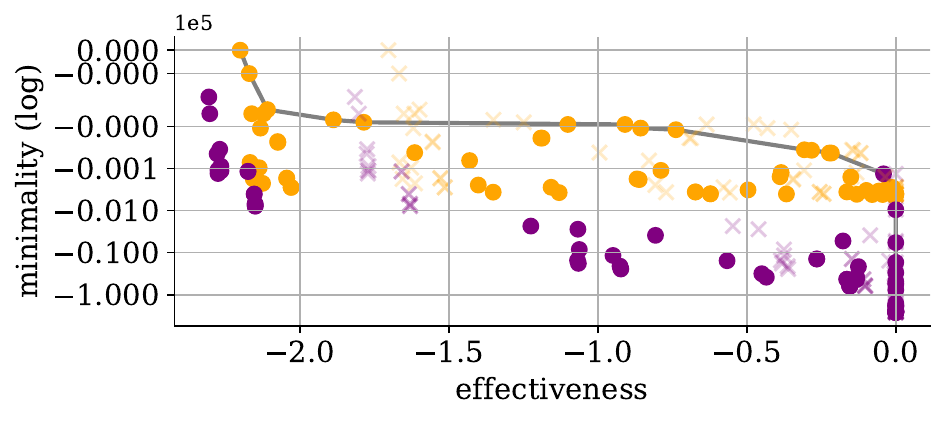}
   \end{subfigure}\vfill
    \raisebox{0.6in}{\rotatebox[origin=t]{90}{\IOI}}  
    \begin{subfigure}[b]{0.48\textwidth}
       \includegraphics[width=\textwidth]{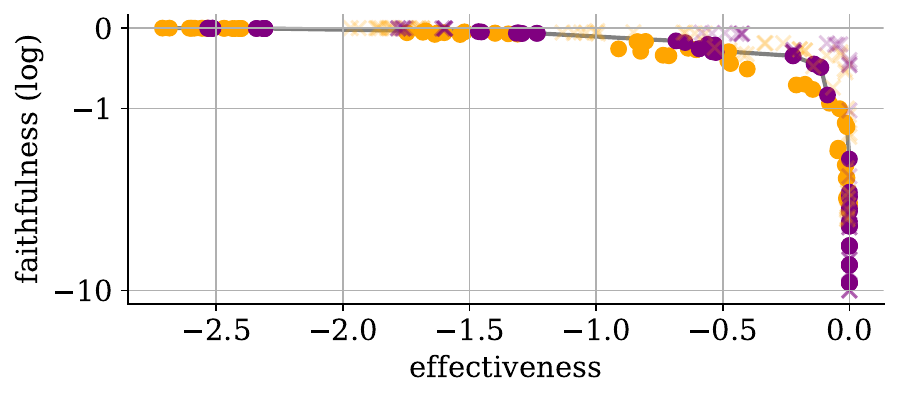}
   \end{subfigure}\hfill
    \begin{subfigure}[b]{0.48\textwidth}
       \includegraphics[width=\textwidth]{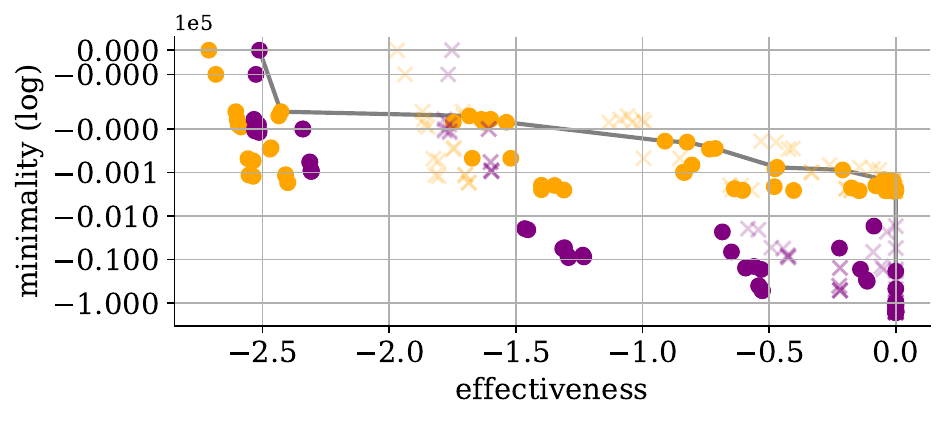}
   \end{subfigure}\hfill
   \caption{Pareto fronts that visualize the trade-off between effectiveness and faithfulness (left) and effectiveness and minimality (right) on train (crosses) and test sets (points). We compare \textcolor{orange}{\FixedAttr} and \textcolor{violet}{\FixedVec} for different hyperparameter combinations of $\lambdaFaith, \lambdaMin, \margin \in \{0, 1, 10, 100\}$. We learn interventions for the sites \attnOut and \mlpOut on all layers and token positions of \gptSmall. We find that \FixedAttr does not fall behind \FixedVec in terms of effectiveness and faithfulness, but is much more minimal on average.}
\label{fig:pareto}
\end{figure*}
 
We intervene on the model to steer its prediction on a selected \defn{task} with data points $\task = \{(\prompt, \correctAnswTok, \wrongAnswTok)\}_{\data=1}^{\Data}$. For each data point, the model is prompted by $\prompt$ to choose between two competing \defn{answer tokens} $\correctAnswTok, \wrongAnswTok \in \Sigma$. The answer tokens are selected such that $\correctAnswTok$ and $\wrongAnswTok$ always represent correct and wrong continuations of the prompt, respectively. For instance, given the prompt $\prompt$ = \promptStr{The capital of Poland is London. Q: What is the capital of Poland? A:}, the competing answer tokens could be $\correctAnswTok = \correctAnsw{Warsaw}$ versus $\wrongAnswTok = \wrongAnsw{London}$. 

\paragraph{Effectiveness.}

A popular objective for learning interventions is the \defn{logit difference} between candidate tokens: $\intervmodel(\prompt)_{\correctAnswTok} - \intervmodel(\prompt)_{\wrongAnswTok}$. We extend this objective to allow the sign of $\beta$ to control the sign of the logit difference. Concretely, we want the logit of $\correctAnswTok$ to be larger than that of $\wrongAnswTok$ by some margin $\margin \geq 0$ when setting $\beta > 0$ and smaller for $\beta < 0$. Define $\intervmodelPlus(\prompt)$ to be the intervention with $\beta = 1$, and $\intervmodelMinus(\prompt)$ to be the intervention with $\beta = {-}1$. The following objective then encourages learned interventions to yield both $\intervmodelPlus(\prompt)_{\correctAnswTok} > \intervmodelPlus(\prompt)_{\wrongAnswTok}$ and $\intervmodelMinus(\prompt)_{\correctAnswTok} < \intervmodelMinus(\prompt)_{\wrongAnswTok}$:
\begin{align}    
    \label{eq:effectiveness}
    \effect{\margin}(\Params, \task) &= \\
    - \frac{1}{\Data} \sum_{\data=1}^{\Data} \big[
    &\max \big(0, \intervmodelPlus(\prompt)_{\wrongAnswTok} {-} \intervmodelPlus(\prompt)_{\correctAnswTok} {+} \margin \big) \nonumber\\ 
     + &\max \big( 0, \intervmodelMinus(\prompt)_{\correctAnswTok} - \intervmodelMinus(\prompt)_{\wrongAnswTok} + \margin \big)  \big]. \nonumber
\end{align}
\paragraph{Faithfulness.}

We say an intervention is faithful if it only affects the answer tokens \citep{wang_interpretability_2023, hanna_have_2024}. We promote faithfulness via the following objective
\begin{align}
\label{eq:faithfulness}
\faith(&\Params, \task)  = \\
 &\quad -\frac{1}{\Data} \sum_{\data=1}^{\Data}  D_{\KL} \Big(\sigma \big(\intervmodelPlus(\prompt) \big)  \mid\mid \sigma \big( \model(\prompt) \big) \Big) \nonumber \\
    &\quad \qquad +  D_{\KL} \Big(\sigma \big(\intervmodelMinus(\prompt) \big)  \mid\mid \sigma \big( \model(\prompt) \big) \Big). \nonumber
\end{align}
where $D_{\KL}$ is the Kullback--Leibler divergence.

\paragraph{Minimality.}
Thirdly, the intervention should be minimal \citep{wang_interpretability_2023}, which we promote with the following regularizing term
\begin{align}
    \minim{\pnorm}(\Params) = -\lVert \mathrm{vec}(\Params)\rVert_{p}
    \label{eq:minimality}
\end{align}
which penalizes a (pseudo)norm of the parameters.
Here, $\mathrm{vec}$ maps the set of parameters $\Params$ to a vector, and the subscript $p$ indicates which (pseudo)norm of the vector is penalized. Setting $p=0$ corresponds to $\ell_0$-regularization, which encourages sparsity directly but is difficult to optimize. We instead take $p=1$, which corresponds to $\ell_1$-regularization, a widely-studied and effective convex relaxation of $\ell_0$-regularization, which forms the basis of the sparsity-inducing LASSO method~\citep{tibshirani_regression_1996}.~\looseness=-1

\paragraph{Gradient-based Parameter Learning.}

Putting it all together, we directly optimize for an intervention that is simultaneously effective, faithful, and minimal.
Specifically, we choose intervention parameters $\Params$ using gradient-based optimization on the multi-term objective 
\begin{equation}
\begin{aligned}
    \label{eq:objective}
    \Psi(\Params, &\task) = \\ &\effect{\margin}(\Params, \task) + \lambdaFaith \faith (\Params, \task) + \lambdaMin \minim{1}(\Params).
\end{aligned}
\end{equation}
We can tune the hyperparameters $\lambdaFaith \geq 0$ and $\lambdaMin \geq 0$ to control the degree to which the three terms in the objective trade off. The margin $m \geq 0 $ constitutes a third hyperparameter that controls the strength of the effectiveness term. 

\subsection{Evaluating an Intervention}
\label{sec:eval}

We can evaluate an intervention based on our operationalizations of effectiveness, faithfulness and minimality. To evaluate effectiveness, we set the margin $\margin = 0$ to obtain a metric ranging from $-\infty$ to 0, where $\effect{0}(\Params, \task) = 0$ indicates that an intervention always successfully flips the answer tokens. The faithfulness objective, which also ranges from $-\infty$ to 0, can be treated as an evaluation metric without any modification. Finally, to evaluate minimality, we count the number of non-negligible intervention parameters---i.e., those taking values sufficiently far from 0. In practice, we consider absolute values less than 0.01 to be negligible.~\looseness=-1

\section{Experiments}
\label{sec:experiments}

We fit the two intervention methods, \FixedAttr and \FixedVec, on the three-term objective in \cref{eq:objective} and conduct evaluations for effectiveness, faithfulness and minimality. 

\subsection{Tasks}
\label{sec:tasks}

We consider two synthetic tasks presented in \cref{tab:datasets}. The Country--Capital Conflicts (\CCC) task, designed by \citet{du_context_2024}, prompts models to resolve an entity-based knowledge conflict~\citep{longpre_entity-based_2021} which pits information provided in-context against prior parametric knowledge that models can be assumed to have acquired during training. The Indirect Object Identification (\IOI) task, which we adapt slightly from \citet{wang_interpretability_2023}, prompts models to choose which of two tokens is the indirect object in sentences with potentially complex syntactic structure. For both tasks, we select prompts to be of the same length $\TokenPos$ and ensure the candidate answers are single tokens.~\looseness=-1

\begin{figure}[t]
 \centering
    \begin{subfigure}[b]{0.24\textwidth}
       \makebox[\textwidth][c]{}
       \includegraphics[width=\textwidth]{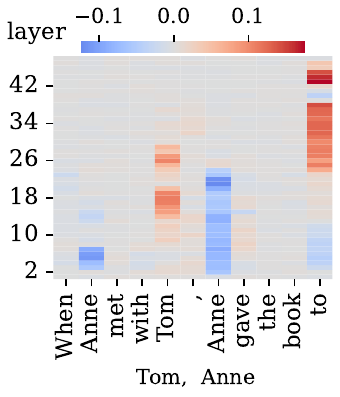}
   \end{subfigure}
   \hfill
    \begin{subfigure}[b]{0.23\textwidth}
       \makebox[\textwidth][c]{}
       \includegraphics[width=\textwidth]{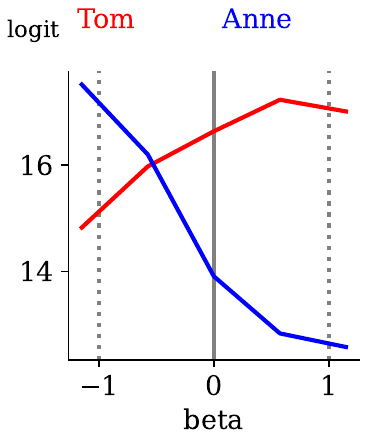}
   \end{subfigure}
   \caption{The magnitude and sign of learned activation scalars highlight task-relevant locations within the model, here for the \residPost site on \gptXL. For instance, the correct answer token \correctAnsw{Tom} is promoted around layers \num{15} to \num{20} while activation vectors at the token \wrongAnsw{Anne} are scaled down. The intervention is successful according to our effectiveness objective illustrated by the red and blue lines crossing between $\beta={-}1$ and $\beta=1$, which reverses the sign of the logit difference.}
\label{fig:interpret_scalars}
\end{figure}

\begin{figure}[t]
 \centering
    \vspace{-0.1cm}
    \begin{subfigure}[b]{0.24\textwidth} 
    \makebox[\textwidth][c]{\footnotesize Abs. change in norm}
    \makebox[\textwidth][c]{\footnotesize corr. over $5$ runs, $\rho = 0.979$}
     \includegraphics[width=\textwidth]{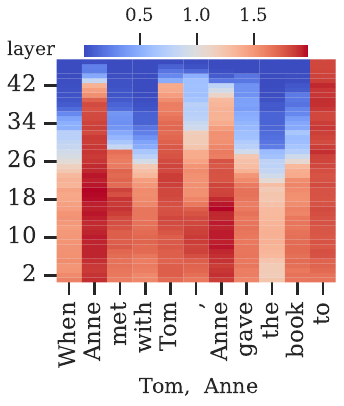}
   \end{subfigure}\hfill
    \begin{subfigure}[b]{0.24\textwidth}
    \makebox[\textwidth][c]{\footnotesize Change in direction (cosine dist)}
    \makebox[\textwidth][c]{\footnotesize corr. over $5$ runs: $\rho = 0.350$}
    \includegraphics[width=\textwidth]{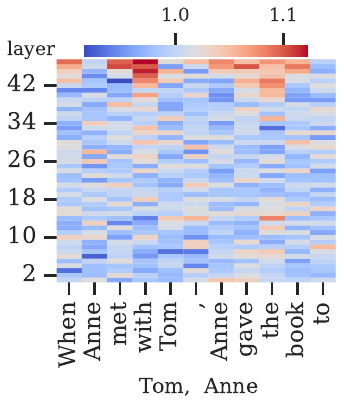}
   \end{subfigure}
   \vfill
\vspace{-0.1cm}
\caption{We interpret the learned steering vectors by comparing their difference in norm and direction, before and after training, here for the \residPost site of \gptXL. We observe that the change in norm highlights similar token positions as \FixedAttr in \cref{fig:interpret_scalars}. When fitting the vectors multiple times, the vectors converge to different directions every time, while changes in norm are more constant. We quantify this using the Kendall correlation $\rho$ between the orderings of locations by the change in norm as well as cosine distance between different runs.}
\label{fig:norm_cosine}
\end{figure}

\subsection{Quantitative Results}
\label{sec:results}
We do not expect there to be a single solution which is optimal for all three objectives. An optimally effective intervention might not be very faithful, while a highly minimal intervention but not be very effective. We seek to understand the trade-off between the different intervention desiderata by finding Pareto-optimal solutions. To this end, we run a grid search over the hyperparameters $\lambdaFaith$, $\lambdaMin$ and $\margin$. We evaluate the learned interventions on the test set and visualize the Pareto frontier in \cref{fig:pareto}.

\paragraph{Effectiveness versus Faithfulness.}

We find that \FixedAttr and \FixedVec perform comparably when it comes to trading off effectiveness and faithfulness. Of all points on the Pareto frontier, around half come from each approach. \FixedVec is generally more effective on the train set, but also exhibits large decreases in test set performance; this is likely due to it having many more parameters and being more prone to overfitting. We also see that high values of effectiveness on the train set are associated with low values of faithfulness on the test set, also suggestive of overfitting.

\paragraph{Effectiveness versus Minimality.}
We find that \FixedAttr is generally more minimal on multiple levels. For \CCC and \IOI, \FixedAttr accounts for about \num{85}\% of the points on the effectiveness-minimality Pareto frontier in \cref{fig:pareto}. Thus, \FixedAttr learns interventions that are in fact more parsimonious. This is in addition to having far fewer learnable parameters. As a simple illustration, consider learning an intervention on the $\Layers = 48$ layers of \gptXL, for a prompt consisting of $\TokenPos = 19$ tokens. If we only intervene on a single site per layer---e.g., \residPost, where the activation vector has dimensionality $D = 1600$---then \FixedVec has $19 \times 48 \times 1600 = \textrm{1,459,200}$ learnable parameters while \FixedAttr has only $48 \times 19 = 912$ learnable scalars. \FixedAttr is also more minimal qualitatively, as it is limited to affecting only the signed magnitude of the activation vectors without otherwise affecting their direction.\looseness=-1

\begin{figure}[t]
 \centering
    \begin{subfigure}[b]{0.235\textwidth}
       \makebox[\textwidth][c]{}
       \includegraphics[width=\textwidth]{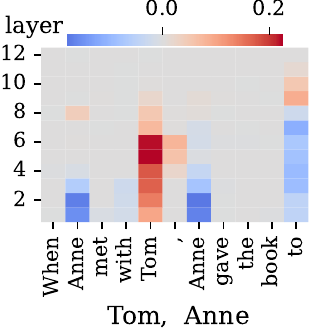}
   \end{subfigure}
   \hfill
    \begin{subfigure}[b]{0.23\textwidth}
       \makebox[\textwidth][c]{}
       \includegraphics[width=\textwidth]{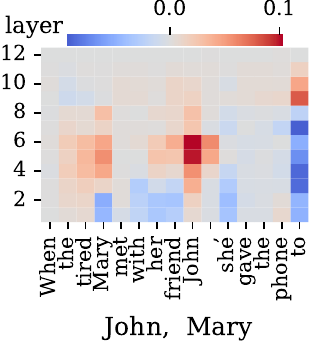}
   \end{subfigure}
   \caption{\DynamicAttr generalizes to prompts of varying length and yields interpretable activation scalars. Here for \residPost of \gptSmall, the correct answer token is identified despite being in a different position.}
\label{fig:dynamicattr_interpr}
\end{figure}

\begin{figure}[ht]
 \centering
    \begin{subfigure}[b]{0.5\textwidth}
       \includegraphics[width=\textwidth]{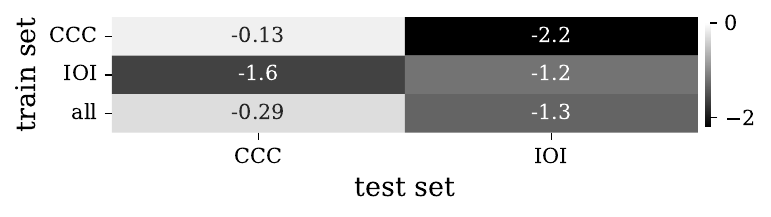}
   \end{subfigure}
 \vspace{-0.8cm}
   \caption{Generalization performance in terms of effectiveness of \DynamicAttr for varying-length prompts from different tasks fitted on \mlpOut and \attnOut of \gptSmall (closer to 0 is better). As expected, training and testing on the same task performs best overall, but training on all tasks generalizes comparatively well.}
\label{fig:dynamicattr_eval}
\end{figure}

\begin{table}[t]
\fontsize{10}{10}\selectfont
\centering
\renewcommand{\arraystretch}{1.2} 
\setlength{\tabcolsep}{0.35em} 
\begin{tabular}{lccc}
\toprule
$\effect{0}$ on \textbf{\CCC}& \textbf{A: train} & \textbf{B: test} & \textbf{C: templ} \\ \midrule
\FixedAttr             &     -0.99   &     -1.57  &  -2.66  \\
\FixedVec              &     -0.06    &     -0.12   &  -2.16  \\
\DynamicAttr           &     -0.11   &     -0.30    &  -2.10
\end{tabular}
\caption{We test generalization performance by learning interventions on \CCC and evaluating effectiveness on prompts of the \textbf{A:} \CCC train set; \textbf{B:} \CCC test set; \textbf{C:} a different template for the \CCC task. All intervention parameters are learned for \attnOut and \mlpOut of \gptSmall, using the same set of hyperparameters.}
\label{tab:ccc_generalization}
\end{table}

\subsection{Interpretation of Scalars and Vectors}

\paragraph{Activation Scalars.}

As shown in \cref{fig:interpret_scalars}, activation scalars highlight task-relevant locations while performing effective interventions at the same time. The norm of activation vectors at the correct in-context tokens is increased, while it is decreased at the incorrect in-context tokens. 

\paragraph{Steering Vectors.}

We seek to better understand the properties of the learned steering vectors by analyzing their change in norm and direction in terms of cosine distance before and after training. As visualized in \cref{fig:norm_cosine}, the change in norm reveals a structured, interpretable pattern, while the directional change appears more arbitrary. In fact, when fitting the steering vectors five times on the same prompt, we can show quantitatively that vectors converge to different solutions in terms of direction but less so in terms of norm.

\section{Extension to Variable-Length Prompts}
\label{sec:unmatched}

\FixedAttr and \FixedVec, as well as most existing interpretability methods such as activation patching, learn intervention parameters that are tied to specific token positions $\tokenPos$ in the prompt $\prompt$. To reuse learned parameters for intervention with a new prompt, the test prompt must match the length $\TokenPos$ of the training prompt, and ideally match it syntactically. Existing work circumvents the requirement of matched prompts by intervening only on the last token position \citep{yu_characterizing_2023, li_inference-time_2023, jin_cutting_2024, stoehr_unsupervised_2024}, or on token positions that can be easily aligned across prompts, like the main verb, or the last token of the subject~\citep{meng_locating_2022, geva_dissecting_2023, ortu_competition_2024, merullo_circuit_2024}.\looseness=-1

\begin{figure}[t!]
 \centering
    \begin{subfigure}[b]{0.24\textwidth}
       \makebox[\textwidth][c]{\footnotesize \FixedAttr on \outVec}
       \includegraphics[width=\textwidth]{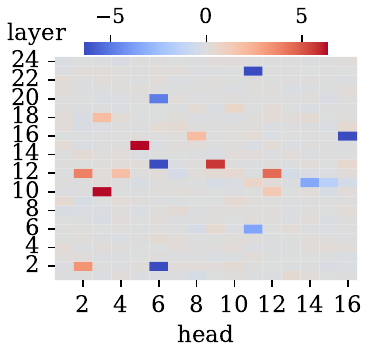}
   \end{subfigure}\hfill
    \begin{subfigure}[b]{0.24\textwidth}
    \makebox[\textwidth][c]{\footnotesize \DLA on \outVec \citep{yu_characterizing_2023}}
       \includegraphics[width=\textwidth]{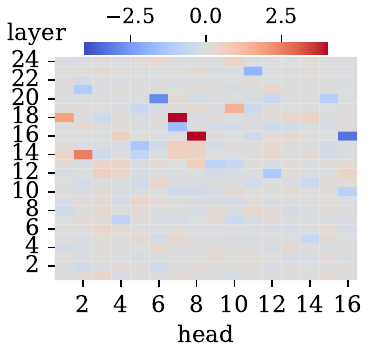}
   \end{subfigure}
   \vfill
    \begin{subfigure}[b]{0.24\textwidth}
       \includegraphics[width=\textwidth]{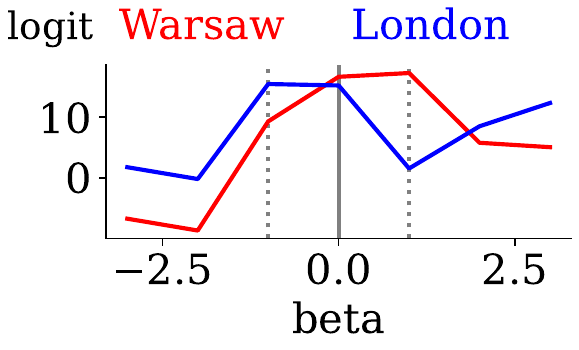}
   \end{subfigure}\hfill
    \begin{subfigure}[b]{0.24\textwidth}
       \includegraphics[width=\textwidth]{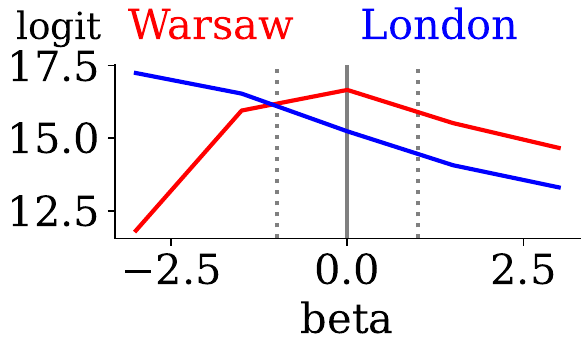}
   \end{subfigure}   \vfill
    \begin{subfigure}[b]{0.24\textwidth}
       \includegraphics[width=\textwidth]{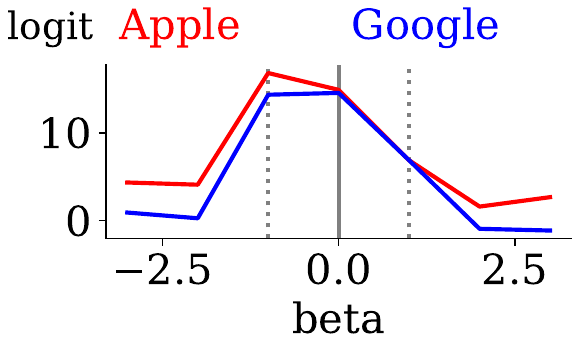}
   \end{subfigure}\hfill
    \begin{subfigure}[b]{0.24\textwidth}
       \includegraphics[width=\textwidth]{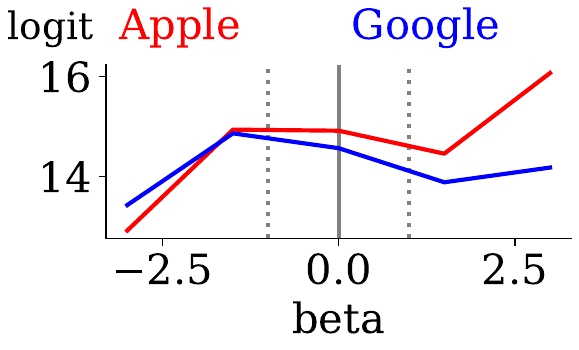}
    \end{subfigure} 
 \vspace{-0.8cm}
\caption{We follow \citet{yu_characterizing_2023} in their approach to identify a memory head in layer $\layer = 16$, head $\head = 8$ in \texttt{Pythia-1.4b} based on Direct Logit Attribution (\DLA) on the last token position. Analogously, we learn activation scalars $\scalar{c}$ for the last token position using \FixedAttr and observe considerable overlap in the attribution scores. Overall, intervening on the last token position appears to be highly effective.}
\label{fig:yu_case_study}
\vspace{-0.5cm}
\end{figure}

\begin{figure}[ht]
 \centering
    \begin{subfigure}[b]{0.24\textwidth}
        \makebox[\textwidth][c]{\footnotesize \FixedAttr on \outVec}
       \includegraphics[width=\textwidth]{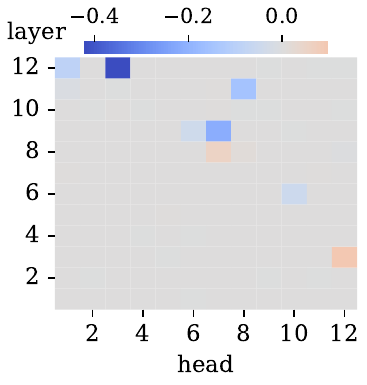}
   \end{subfigure}\hfill
    \begin{subfigure}[b]{0.24\textwidth}
           \makebox[\textwidth][c]{\footnotesize \citet{ortu_competition_2024} on \outVec}
       \includegraphics[width=\textwidth]{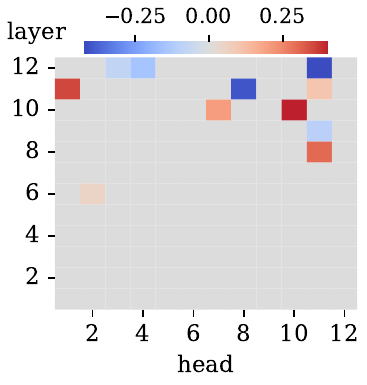}
   \end{subfigure}
   \vfill
    \begin{subfigure}[b]{0.24\textwidth}
       \includegraphics[width=\textwidth]{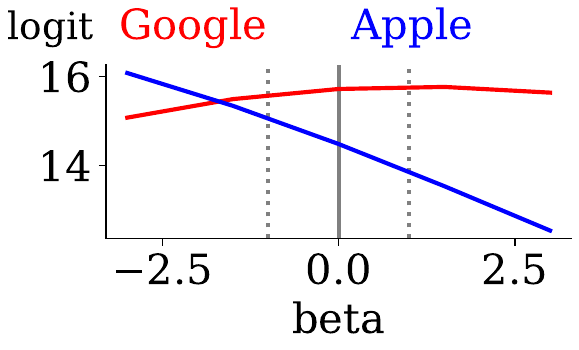}
   \end{subfigure}\hfill
    \begin{subfigure}[b]{0.24\textwidth}
       \includegraphics[width=\textwidth]{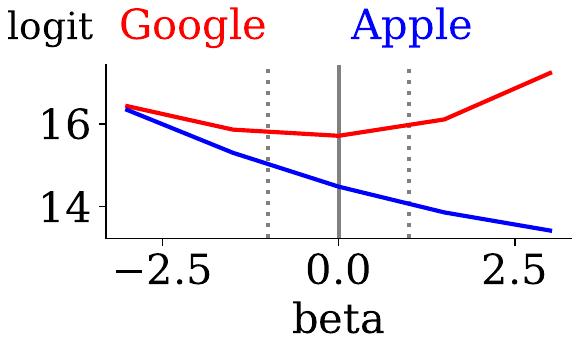}
   \end{subfigure}       
   \caption{We compare \FixedAttr on \gptSmall with a loosely replicated version of the attribution scores in Fig. 4(a) in \citet{ortu_competition_2024}. In our version, we study interventions on the \outVec activation vectors at the last token position. Given the prompt \promptStr{iPhone was developed by Google. iPhone was developed by},  \{\correctAnsw{Google}, \wrongAnsw{Apple}\}, both approaches single out similar attention heads, e.g., layer $\layer = 11$, head $\head = 8$.
   }
\label{fig:ortu_case_study}
\end{figure}

\begin{figure*}[t!]
 \centering
    \begin{subfigure}[b]{0.25\textwidth}
       \makebox[\textwidth][c]{\footnotesize \ActivPatch}
       \includegraphics[width=\textwidth]{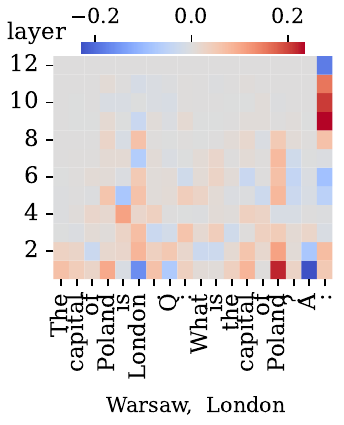}
   \end{subfigure}\hfill
    \begin{subfigure}[b]{0.25\textwidth}
        \makebox[\textwidth][c]{\footnotesize \AttrPatch}
       \includegraphics[width=\textwidth]{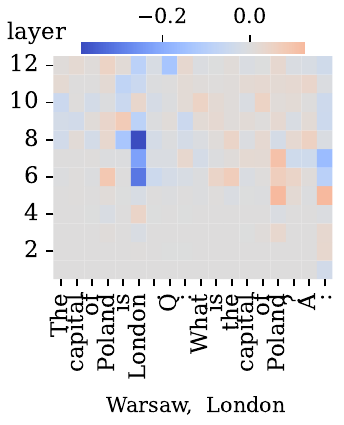}
   \end{subfigure}\hfill
       \begin{subfigure}[b]{0.25\textwidth}
       \makebox[\textwidth][c]{\footnotesize \DLA}
       \includegraphics[width=\textwidth]{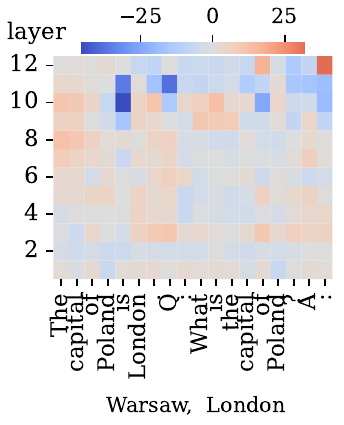}
   \end{subfigure}\hfill
       \begin{subfigure}[b]{0.25\textwidth}
       \makebox[\textwidth][c]{\footnotesize \FixedAttr}
       \includegraphics[width=\textwidth]{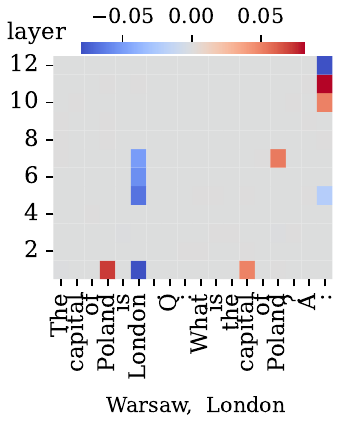}
   \end{subfigure}\hfill
   \vfill
    \begin{subfigure}[b]{0.25\textwidth}
       \includegraphics[width=\textwidth]{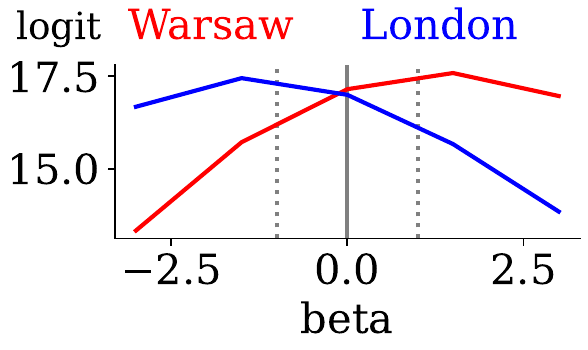}
   \end{subfigure}\hfill
    \begin{subfigure}[b]{0.25\textwidth}
       \includegraphics[width=\textwidth]{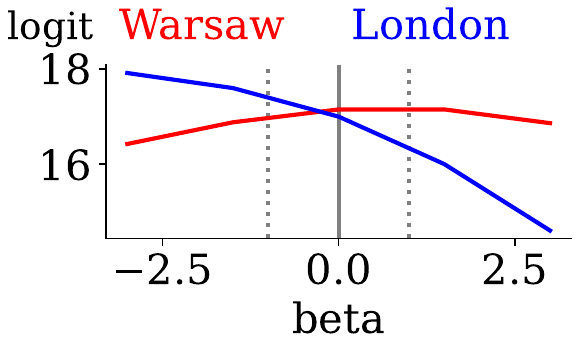}
   \end{subfigure}\hfill
       \begin{subfigure}[b]{0.25\textwidth}
       \includegraphics[width=\textwidth]{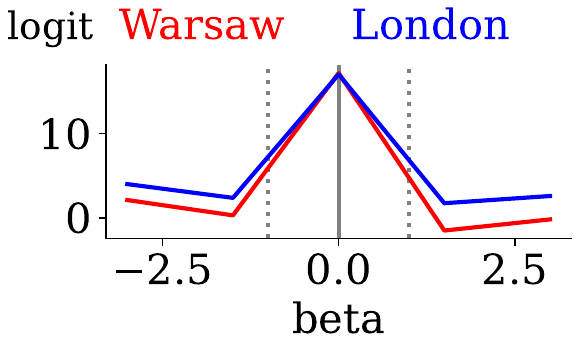}
   \end{subfigure}\hfill
       \begin{subfigure}[b]{0.25\textwidth}
       \includegraphics[width=\textwidth]{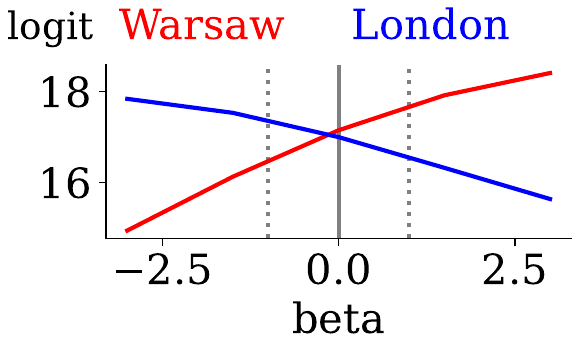}
   \end{subfigure}\hfill
   \vfill
    \begin{subfigure}[b]{0.25\textwidth}
       \includegraphics[width=\textwidth]{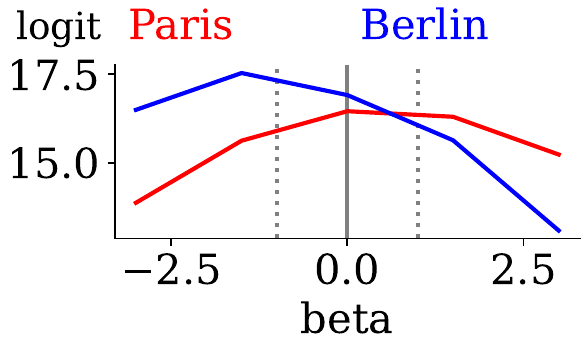}
   \end{subfigure}\hfill
    \begin{subfigure}[b]{0.25\textwidth}
       \includegraphics[width=\textwidth]{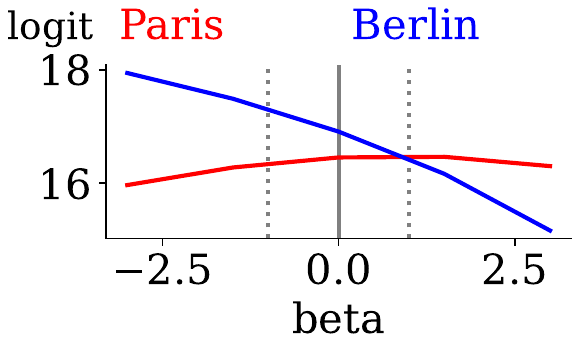}
   \end{subfigure}\hfill
       \begin{subfigure}[b]{0.25\textwidth}
       \includegraphics[width=\textwidth]{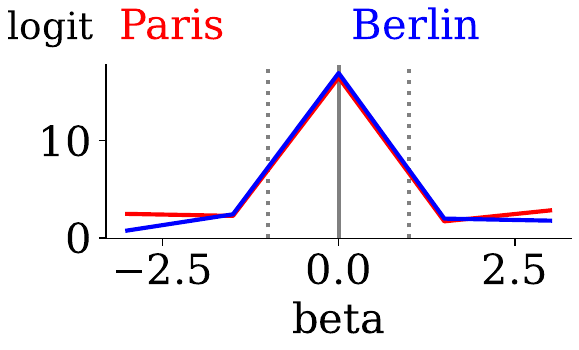}
   \end{subfigure}\hfill
       \begin{subfigure}[b]{0.25\textwidth}
       \includegraphics[width=\textwidth]{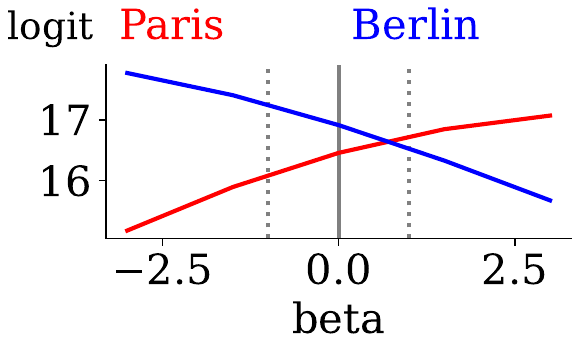}
   \end{subfigure}\hfill
   \caption{\textbf{[Top]} We fit a variant of \ActivPatch, \AttrPatch, \DLA, and \FixedAttr (left to right) on the \mlpOut and \attnOut sites of \gptSmall. We find that the highlighted locations overlap, across methods, at important tokens such as \wrongAnsw{London}. \textbf{[Middle]} Intervening on the training prompt is effective for all methods except \DLA. \textbf{[Bottom]} All methods (except \DLA) generalize to a test set prompt of a different country--capital conflict.}
\label{fig:warsaw_london}
\end{figure*}

\paragraph{Dynamic Activation Scalars.}

We propose an extension of \FixedAttr which defines each scalar to be a function of the corresponding activation vector. Each function is tied to a particular layer $\layer$ and site $\site$, but shared across token positions $\tokenPos$, and parameterized by a (column) vector $\sfunc \in \mathbb{R}^{D}$. The set of learnable intervention parameters is thus $ \theta = \{ \sfunc \}_{(\layer, \site) \in \layerSet \times \siteSet}$. The scalar $\scalar{(\site)}$ for a given position $\tokenPos$ is then the following function of the corresponding activation vector:
\begin{align}
    \scalar{(\site)} &\defequals {\sfunc}^\top  \left(\tfrac{\activs{\layer, \tokenPos}{(\site)}}{\lVert\activs{\layer, \tokenPos}{(\site)} \rVert_2}\right).
    \label{eq:dynamic_scalars}
\end{align}
When combined with~\cref{eq:scalars}, this defines a dynamic intervention (\DynamicAttr). Learning this intervention can be understood as learning probes that identify task-relevant activation vectors and then strengthen or weaken their magnitude via multiplicative scalars. In \cref{fig:dynamicattr_interpr}, we show how \DynamicAttr generalizes to prompts of varying length while still offering interpretable insights highlighting task-relevant tokens. In \cref{fig:dynamicattr_eval}, we evaluate the generalization performance of \DynamicAttr in terms of effectiveness. As expected, performance is highest when test sets are of the same task as the train set; however, \DynamicAttr also obtains good results when trained on a mix of tasks.\looseness=-1

In \cref{tab:ccc_generalization}, we quantify how well \FixedAttr, \FixedVec and \DynamicAttr generalize to varying degrees of domain shift when trained on the \CCC task. All prompts are of the same length in order for \FixedAttr and \FixedVec to be applicable, but the prompt template and thus the position $\tokenPos$ of task-relevant tokens changes. For the selected hyperparameter setting, we find that \DynamicAttr and \FixedVec show better generalization performance than \FixedAttr. The strong generalization of \FixedVec is surprising given that the learned steering vectors are tied to specific token positions. We hypothesize that much of the steering performance should be attributed to the information processed at the last token position \citep{yu_characterizing_2023, wu_language_2024, ortu_competition_2024}, which is corroborated by the heatmap in \cref{fig:interpret_scalars}. We further explore this hypothesis in the next section.

\paragraph{Scaling at the Last Token Position.}

\FixedAttr and \FixedVec show high steering effectiveness even when the prompt template is changed. This suggests a gate-keeping role of computations at the last token position which suppresses or promotes task-relevant information. \citet{yu_characterizing_2023} rely on direct logit attribution \citep[\DLA;][]{elhage_mathematical_2021} on the output activation vector (\outVec) at the last token position.\footnote{Please refer to \cref{sec:attention} for details on the attention mechanism and our naming conventions of activation vectors.} They identify what they call a memory head in layer $\layer = 16$, head $\head = 8$ of \pythiaMedium. They then intervene on this head by scaling the value activation vector (\vVec). We replicate a version of their method and contrast it with \FixedAttr in \cref{fig:yu_case_study}. We find high attribution scores for similar heads, but also single out heads in lower layers, which is less typical for \DLA. We conduct another case study on the role of the last token position comparing \FixedAttr with a version of \citet{ortu_competition_2024}, presented in \cref{fig:ortu_case_study}.\looseness=-1

\section{Related Work}
\label{sec:related_work}

\subsection{Repurposing Interpretability Methods}
\label{sec:repurpose}

There exist various methods producing activation-level attribution scores based off a logit difference metric between two answer tokens. Activation patching \citep[\ActivPatch;][]{lakretz_emergence_2019, vig_causal_2020, meng_locating_2022} swaps activation vectors in one forward pass with activation vectors from another (corrupted) forward pass to study (causal) effects on the logit difference. Attribution patching
\citep[\AttrPatch;][]{nanda_attribution_2023, syed_attribution_2023, kramar_atp_2024} represents a gradient-based, efficient approximation of \ActivPatch. Direct logit attribution \citep[\DLA;][]{elhage_mathematical_2021} identifies activation vectors whose directions are correlated with the vectors of the answer tokens in the projection matrix.\looseness=-1

All three methods produces attribution scores for individual model components, but whether these scores are informative for model-wide steering is unclear. Our work offers a convenient framework to study this question. We can repurpose attribution scores obtained with \ActivPatch, \AttrPatch and \DLA and treat them as activation scalars $\scalar{(s)}$ by plugging them directly into \cref{eq:scalars}. Since these scores were not specifically trained for our task, we can then tune the $\beta$ parameter to globally strengthen, weaken or flip the interventional effect as presented in \cref{fig:warsaw_london}.

For the given prompt, we find that \ActivPatch, and \AttrPatch facilitate successful steering according to our effectiveness metric \cref{eq:effectiveness}. This is surprising as these methods are not trained on this objective and their interventional strength may be on a different scale, i.e., not flipping the answer tokens between $\beta = -1$ and $\beta = 1$. In \cref{fig:warsaw_london} for instance, we find that attribution scores from \DLA are on a much larger scale. This is pointing at a conceptual difference between \FixedAttr and existing methods: \FixedAttr is trained on an explicit objective serving steerability purposes, thus learning an intervention with a well-defined scale and steering direction. Besides, gradient-based learning is faster than activation patching in most cases and incorporates inter-dependencies between multiple activation vectors.

\subsection{Other Related Work}

\paragraph{Transformer Circuits.}

Joint interpretability and steerability is also a desired property in circuit discovery that seeks to identify a minimal subgraph responsible for the behavior of the full model when solving a specific task \citep{wang_interpretability_2023, bhaskar_finding_2024}. Isolating the subgraph typically requires thresholding the attribution scores associated with model components and then zeroing out \citep{de_cao_sparse_2022, wang_interpretability_2023, conmy_towards_2023, syed_attribution_2023} or corrupting them \citep{geiger_causal_2021, bhaskar_finding_2024}. However, this means that the intervention can be considered discrete, as it cannot smoothly facilitate different intervention strengths and directions. The scaling aspect of the $\beta$ hyperparameters in \FixedAttr and \FixedVec, in turn, is continuous. 

\paragraph{Gradient-Based Steering and Interpretability.}

This work relies on gradient-based optimization to localize model components that are relevant with respect to a specifically designed objective. This is similar to \citet{subramani_extracting_2022, hernandez_inspecting_2024}
that learn steering vectors to edit activation vectors. Other related work analyzes the direction and magnitude of weight or activation gradients \citep{du_generalizing_2023, stoehr_localizing_2024, katz_backward_2024} to identify task-relevant components.\looseness=-1

\paragraph{Pruning, Masking and Adapters.}

Learning activation scalars bears resemblance to work on pruning neural networks \citep{li_differentiable_2021}, fine-tuning (low-rank) adapters \citep{houlsby_parameter-efficient_2019, hu_lora_2021} and (hard) masking \citep{louizos_learning_2018, bondarenko_quantizable_2023}. In this work, however, we do not pursue the typical goals of the pruning literature, which often focuses on reducing the computational cost of inference. Instead, \FixedAttr can be seen as learning a \emph{soft} mask that strengthens or weakens components for the purpose of obtaining an interpretable map of locations.

\section{Conclusion}
\label{sec:conclusion}

We show that scaling the signed magnitude of a few relevant activation vectors is often sufficient to flip a model's prediction between a correct and a wrong answer token. Besides being effective at steering, activation scaling requires many fewer parameters than additive steering vectors which intervene both on the magnitude and direction of activation vectors. Our gradient-based multi-objective learning scheme can be understood as reversing the interpretability pipeline, putting steering performance based on clearly defined objectives first and interpretability as a natural by-product second. 

\section*{Acknowledgments}

We thank Cl\'ement Dumas and Alessandro Stolfo for helpful early-stage discussions. Niklas Stoehr is supported by the Swiss Data Science Center (SDSC) PhD fellowship. Vésteinn Snæbjarnarson is funded by the Pioneer Centre for AI, DNRF grant number P1.

\section*{Limitations}
\label{sec:limitations}

\FixedVec, \FixedAttr and \DynamicAttr are controllable via different hyperparameters: the margin $\margin$ in the effectiveness objective, $\lambdaFaith$ weighing the faithfulness term and $\lambdaMin$ weighing the strength of the $\ell_1$-regularization. There are additional training-related hyperparameters such as the learning rate, the number of epochs, the batch size, the number of data instances, and the standard deviation of the Gaussian noise initialization of the intervention parameters, that have a strong influence on the results. For instance, increasing the noise or the margin, training for more epochs, or weakening the $\ell_1$-regularization, results in activation scalars that deviate more from zero.

Access to many hyperparameters can make a method more difficult to deploy. On the other hand, hyperparameters with well-founded semantic can offer desirable, fine-grained controls. For instance, a larger $\lambdaMin$ hyperparameter leads to sparser activation scalars that are easier to interpret, a level of control not offered by many existing methods. Yet, methods like \ActivPatch or \AttrPatch also require finicky hyperparameter choices such as how to corrupt the prompt, e.g., deciding on the standard deviation of the Gaussian noise \citep{meng_locating_2022} to obtain a second corrupted prompt for patching.\looseness=-1

A limitation of this work is the size of models studied and the small size and synthetic character of the tasks. The two largest models considered in this work are \gptXL (1.5 billion parameters) and \pythiaMedium (1.4 billion parameters). Beyond the required compute, we do not anticipate problems applying activation scaling to larger models. Testing \FixedAttr on more real-world datasets with longer prompts is another future avenue. To boost the performance of \DynamicAttr, one could expand the computational expressivity of the activation vector-to-scalar function $\sfunc$. Finally, more work is needed in extending our evaluation based on effectiveness to existing methods such as activation patching. For instance, a promising direction is to fix the attribution scores obtained from activation patching and then post-hoc learning a suitable $\beta$ parameter that facilities the answer token flipping behavior.\looseness=-1

\section*{Impact Statement}

This work aims to better understand the internal workings of language models. This understanding may serve the post-hoc identification of harmful properties such as hallucination, illicit memorization, and undesired biases. It should ideally help in taking preemptive action to guide the design and training of future models. The required compute to apply activation scaling is predominantly dictated by the size of the studied language models. 

\bibliography{main}

\appendix

\section{Appendix}
\label{sec:appendix}

\subsection{Technical Details}
\label{sec:technical}

We implement all steering and interpretability methods using \href{https://transformerlensorg.github.io/TransformerLens/}{\texttt{TransformerLens}} \citep{nanda_transformerlenslibrary_2023} and hyperparameters are chosen based on a combination of the grid search displayed in \cref{fig:pareto} as well as steering and interpretability desiderata specific to each setting. We train gradient-based methods using the Adam optimizer \citep{kingma_adam_2015} for \num{25} epochs. We typically choose a smaller learning rate of \num{0.0001} for training \FixedVec and \num{0.001} for \FixedAttr when training on a single data point. A similarly influential hyperparameter is the initialization of the trainable intervention parameters $\Params$ that we initialize with Gaussian noise $\calN(0, 1\mathrm{e}-5)$. \looseness=-1

\subsection{Multi-headed Attention}
\label{sec:attention}

We now describe the multi-headed attention mechanism $\ATTN{l}$ \citep{vaswani_attention_2017} in more detail. Given a $D \times \TokenPos$ activation matrix $\Activs{\layer-1}{(4)} = [\activs{\layer-1,1}{(4)}, \ldots, \activs{\layer-1, \TokenPos}{(4)}]$, the multi-headed attention mechanism at layer $\layer$ computes 
\begin{subequations}
\begin{align}  
\Activs{\layer}{(1)} &\overset{}{=} \ATTN{\layer}\big( \layerNorm{\layer}{(1)}(\Activs{\layer-1}{(4)}) \big)\\
 &\overset{}{=} \sum_{\head=1}^{\Head} \weightMatrix{\head, \layer}{(\beforeOut)} \AttnHead_{\head, l}^{} \big( \layerNorm{\layer}{(1)}(\Activs{\layer-1}{(4)}) \big)\\
 &\overset{}{=} \sum_{\head=1}^{\Head} \weightMatrix{\head, \layer}{(\beforeOut)} \Activs{\head, \layer}{(\beforeOut)}\\
&\overset{}{=} \sum_{\head=1}^{\Head} \Activs{\head, \layer}{(\out)}.
\label{eq:attention} %
\end{align}%
\end{subequations}%
Essentially, each individual \defn{attention head} $\AttnHead_{\head, l}^{}$ with head index $\head \in \{ 1, \ldots, \Head \}$ computes an activation matrix $\Activs{\head, \layer}{(\beforeOut)} \in \mathbb{R}^{D' \times \TokenPos}$. This per-head matrix is then multiplied with $\weightMatrix{\head, \layer}{(\beforeOut)} \in \mathbb{R}^{D \times D'}$ to obtain a per-head output activation matrix $\Activs{\head, \layer}{(\out)} \in \mathbb{R}^{D \times \TokenPos}$. Note that $\activs{\head, \layer, \tokenPos}{(\beforeOut)} \in \mathbb{R}^{D'}$ and $\activs{\head, \layer, \tokenPos}{(\out)} \in \mathbb{R}^{D}$ are column vectors of $\Activs{\head, \layer}{(\beforeOut)}$ and $\Activs{\head, \layer}{(\out)}$, respectively.

To zoom in on individual attention heads, we now omit the head index $\head$ and layer index $\layer$ for notational simplicity. Under this simplified notation, we define $\AttnHead(\Activs{}{})$ as follows. Each activation (column) vector $\activs{\tokenPos}{} \in \mathbb{R}^{D}$ of the matrix $\Activs{}{} \in \mathbb{R}^{D \times \TokenPos}$ is linearly projected to compute query, key and value activation vectors according to
\begin{subequations}
\begin{align}   
\activs{\tokenPos}{(\query)} &\overset{}{=} \weightMatrix{}{(\query)} \activs{\tokenPos}{}\\
\activs{\tokenPos}{(\key)} &\overset{}{=} \weightMatrix{}{(\key)} \activs{\tokenPos}{}\\
\activs{\tokenPos}{(\val)} &\overset{}{=} \weightMatrix{}{(\val)} \activs{\tokenPos}{}
\label{eq:KQV} 
\end{align}%
\end{subequations}%
where $\weightMatrix{}{(\query)}, \weightMatrix{}{(\key)}, \weightMatrix{}{(\val)} \in \mathbb{R}^{D' \times D}$. The key and value vectors are then used to compute $\TokenPos$ different \defn{self-attention distributions} $\boldsymbol{\kappa}_{i}$ \citep{li_transformer_2024} over the probability simplex $\Delta^{\TokenPos-1}$ following
\begin{subequations}
\begin{align}   
\overline{\kappa}_{\tokenPos}(j) &\overset{}{=} \frac{\activs{\tokenPos}{(\query)}^{\top} \activs{j}{(\key)}}{\sqrt{D'}} \\
\boldsymbol{\kappa}_{\tokenPos} &\overset{}{=} \sigma([\overline{\kappa}_{\tokenPos}(1), \ldots, \overline{\kappa}_{\tokenPos}(\TokenPos)]^\top)%
\label{eq:attention_distribution} %
\end{align}%
\end{subequations}%
where $\overline{\kappa}_{\tokenPos}(j)$ represents the (unnormalized) attention score that token position $\tokenPos$ pays to token position $j$ and $\sigma$ is the softmax function. Importantly, in autoregressive language modeling, it is common to apply the hard constraint of an attention mask to disallow token positions earlier in the prompt to attend to positions later in the prompt---i.e., $j \leq i$. 

Finally, the self-attention distributions are used to construct a weighted average of the value vectors $\activs{\tokenPos}{(\val)}$ according to
\begin{subequations}
\begin{align}   
\activs{\tokenPos}{(\beforeOut)}  &\overset{}{=} \sum_{j=1}^{\TokenPos} \kappa_{\tokenPos}(j) \activs{j}{(v)}%
\label{eq:weighted_values}%
\end{align}%
\end{subequations}%
In the main part of this paper, \cref{fig:yu_case_study} and \cref{fig:ortu_case_study} specifically, we refer to $\activs{\head, \layer, \tokenPos}{(\beforeOut)}$ as \zVec, $\activs{\head, \layer, \tokenPos}{(\out)}$ as \outVec and $\activs{\head, \layer, \tokenPos}{(\vVec)}$ as \vVec activation vectors.

\end{document}

%% file: preamble/preamble.tex
\usepackage{microtype}
\usepackage[utf8]{inputenc}
\usepackage[T1]{fontenc}
\usepackage{times}
\usepackage{latexsym}
\usepackage{amsfonts}
\usepackage{amsmath}
\usepackage{amssymb}
\usepackage{amsthm}
\usepackage{relsize}
\usepackage{xspace}
\usepackage{comment}
\usepackage{placeins} 

\newcommand{\setQuote}[1]{``{#1}''}
\newcommand{\defn}[1]{\textbf{#1}}

\usepackage[group-separator={,}]{siunitx}

\usepackage{tikz}
\usetikzlibrary{bayesnet}
\usetikzlibrary{arrows}
\usepackage{color}
\usepackage{graphicx}
\usepackage{caption}
\usepackage{subcaption}
\usetikzlibrary{backgrounds}

\usepackage{hhline}
\usepackage{xcolor}
\usepackage{colortbl}
\definecolor{Grey}{RGB}{210,210,210}

\def\calN{{\mathcal{N}}}

\def\calT{{\mathcal{T}}}

\usepackage{enumitem}

\usepackage{algorithm}
\usepackage{algpseudocode}

\usepackage{multirow}

%% file: preamble/math_commands.tex
\usepackage{amsmath,amsfonts,bm}
\usepackage{dsfont}

\def\eqref#1{equation~\ref{#1}}

\def\1{\bm{1}}

\DeclareMathAlphabet{\mathsfit}{\encodingdefault}{\sfdefault}{m}{sl}
\SetMathAlphabet{\mathsfit}{bold}{\encodingdefault}{\sfdefault}{bx}{n}



%% file: preamble/notation.tex
\usepackage{wasysym}

\definecolor{ETHPetrol}{RGB}{0,120,148}	
\colorlet{MacroColor}{black}
\newcommand{\mymacro}[1]{{\color{MacroColor} #1}}

\newcommand{\promptColor}[1]{{\color{gray} #1}}
\newcommand{\correctColor}[1]{{\color{red} #1}}
\newcommand{\wrongColor}[1]{{\color{blue} #1}}

\newcommand{\MLP}[1]{\mymacro{\text{MLP}_{#1}}}
\newcommand{\ATTN}[1]{\mymacro{\text{ATTN}_{#1}}}

\newcommand{\mlpOut}{\mymacro{\texttt{mlpOut}}\xspace}
\newcommand{\attnOut}{\mymacro{\texttt{attnOut}}\xspace}

\newcommand{\residPost}{\mymacro{\texttt{residPost}}\xspace}

\newcommand{\zVec}{\mymacro{\texttt{z}}\xspace}
\newcommand{\outVec}{\mymacro{\texttt{o}}\xspace}
\newcommand{\vVec}{\mymacro{\texttt{v}}\xspace}

\newcommand{\pythia}{\mymacro{\texttt{Pythia}}\xspace}
\newcommand{\pythiaMedium}{\mymacro{\texttt{Pythia-1.4B}}\xspace}
\newcommand{\gpt}{\mymacro{\texttt{GPT2}}\xspace}
\newcommand{\gptXL}{\mymacro{\texttt{GPT2-XL}}\xspace}
\newcommand{\gptSmall}{\mymacro{\texttt{GPT2-Small}}\xspace}

\newcommand{\ActivPatch}{\mymacro{\textsc{ActivPatch}}\xspace}
\newcommand{\AttrPatch}{\mymacro{\textsc{AttrPatch}}\xspace}
\newcommand{\DLA}{\mymacro{\textsc{DLA}}\xspace}
\newcommand{\FixedAttr}{\textsc{ActivScalar}\xspace}
\newcommand{\FixedVec}{\textsc{SteerVec}\xspace}
\newcommand{\DynamicAttr}{\mymacro{\textsc{DynScalar}}\xspace}

\newcommand{\IOI}{\mymacro{\textsc{IOI}}\xspace}
\newcommand{\CCC}{\mymacro{\textsc{CCC}}\xspace}

\newcommand{\stringtok}{\mymacro{v}}
\newcommand{\String}{\mymacro{\boldsymbol{v}}}

\newcommand{\layer}{\mymacro{l}}
\newcommand{\Layers}{\mymacro{L}}
\newcommand{\layerSet}{\mymacro{\mathcal{L}}}

\newcommand{\tokenPos}{\mymacro{i}}
\newcommand{\TokenPos}{\mymacro{I_{\data}}}
\newcommand{\tokenSet}{\mymacro{\mathcal{I}}}

\newcommand{\site}{\mymacro{s}}

\newcommand{\siteSet}{\mymacro{\mathcal{S}}}
\newcommand{\Points}{\mymacro{\mathcal{K}}}

\newcommand{\head}{\mymacro{t}}
\newcommand{\Head}{\mymacro{T}}
\newcommand{\AttnHead}{\mymacro{A}}
\newcommand{\val}{\mymacro{v}}
\newcommand{\key}{\mymacro{k}}
\newcommand{\query}{\mymacro{q}}
\newcommand{\beforeOut}{\mymacro{z}}
\newcommand{\out}{\mymacro{o}}

\newcommand{\task}{\mymacro{\calT}}
\newcommand{\data}{n}
\newcommand{\Data}{\mymacro{N}}

\newcommand{\promptStr}[1]{\textit{\promptColor{#1}}}
\newcommand{\correctAnsw}[1]{\textit{\correctColor{#1}}}
\newcommand{\wrongAnsw}[1]{\textit{\wrongColor{#1}}}

\newcommand{\correctAnswTok}{\correctColor{c_{\data}}}
\newcommand{\wrongAnswTok}{\wrongColor{w_{\data}}}

\newcommand{\sy}{\mymacro{y_{\data}}}
\newcommand{\prompt}{\promptColor{{\boldsymbol{x}_{\data}}}}

\newcommand{\activs}[2]{\mymacro{\mathbf{h}_{#1}^{{#2}}}}
\newcommand{\Activs}[2]{\mymacro{\mathbf{H}_{#1}^{{#2}}}}
\newcommand{\changedActivs}[2]{\mymacro{\widetilde{\mathbf{h}}_{#1}^{{#2}}}}

\newcommand{\sfunc}{\mymacro{\mathbf{g}_{\layer}^{(\site)}}}
\newcommand{\scalar}[1]{\mymacro{\alpha_{\layer,\tokenPos}^{#1}}}
\newcommand{\vecto}[1]{\mymacro{\mathbf{v}_{\layer,\tokenPos}^{#1}}}

\newcommand{\Params}{\mymacro{\boldsymbol{\theta}}}

\newcommand{\model}{\mymacro{f}}
\newcommand{\margin}{\mymacro{m}}

\newcommand{\intervmodel}{\mymacro{\widetilde{f}_{\Params}^{\beta}}}
\newcommand{\intervmodelPlus}{\mymacro{\widetilde{f}_{\Params}^{+}}}
\newcommand{\intervmodelMinus}{\mymacro{\widetilde{f}_{\Params}^{-}}}

\newcommand{\Unembed}{\mymacro{\mathbf{E}}}

\newcommand{\EOS}{\mymacro{\textsc{eos}}}
\newcommand{\SigmaStar}{\mymacro{\Sigma^{*}}}
\newcommand{\SigmaBar}{\mymacro{\overline{\Sigma}}}

\newcommand{\prob}{\mymacro{p}}
\newcommand{\KL}{\mymacro{\mathrm{KL}}}

\newcommand{\effect}[1]{\mymacro{\textrm{E}_{#1}}}
\newcommand{\faith}{\mymacro{\textrm{F}}}
\newcommand{\minim}[1]{\mymacro{\textrm{M}_{#1}}}

\newcommand{\pnorm}{p}

\newcommand{\lambdaFaith}{\mymacro{\lambda_{\mathsmaller{\textrm{F}}}}}
\newcommand{\lambdaMin}{\mymacro{\lambda_{\mathsmaller{\textrm{M}}}}}

\newcommand{\layerNorm}[2]{\mymacro{\mathrm{LN}_{#1}^{#2}}}
\newcommand{\weightMatrix}[2]{\mymacro{\mathbf{W}_{#1}^{#2}}}

\newcommand{\defequals}{\triangleq}

%% file: preamble/bibstyle.tex
\usepackage{cleveref}
\crefname{section}{\S}{\S\S}
\Crefname{section}{\S}{\S\S}
\crefname{table}{Tab.}{}
\crefname{figure}{Fig.}{}
\crefname{algorithm}{Algorithm}{}
\crefname{equation}{Eq.}{}
\crefname{appendix}{App.}{}
\crefname{thm}{Theorem}{}
\crefname{prop}{Proposition}{}
\crefname{cor}{Corollary}{}
\crefname{observation}{Observation}{}
\crefname{assumption}{Assumption}{}
\crefformat{section}{\S#2#1#3}

\makeatletter
\newcommand\footnoteref[1]{\protected@xdef\@thefnmark{\ref{#1}}\@footnotemark}
\makeatother